%% file: icra.tex
\newif\ifarxivversion
\arxivversiontrue
\documentclass[letterpaper,10pt,conference]{ieeeconf}
\IEEEoverridecommandlockouts
\usepackage[T1]{fontenc}
\usepackage{mathptmx}
\usepackage{tabularx}
\usepackage{graphicx}
\usepackage{amsmath,amssymb}
\usepackage{cuted}
\let\labelindent\relax
\usepackage{enumitem}
\usepackage{booktabs,multirow,array}
\usepackage[table]{xcolor}
\usepackage{cite}

\ifarxivversion 
\makeatletter
\let\NAT@parse\undefined
\makeatother
\else
\fi

\usepackage[
    colorlinks=true,
    linkcolor=blue,
    citecolor=black,
    urlcolor=blue
]{hyperref}
\makeatletter
\let\ICRA@makecaption\@makecaption
\long\def\@makecaption#1#2{%
  \begingroup
    \let\footnotesize\normalsize
    \ifx\@captype\@IEEEtablestring%
      \setbox\@tempboxa\hbox{\footnotesize #1: #2}%
      \ifdim\wd\@tempboxa>\hsize
        \parbox[t]{\hsize}{\footnotesize\noindent #1: #2}%
      \else
        \hbox to\hsize{\footnotesize\hfil\box\@tempboxa\hfil}%
      \fi
      \@IEEEtablecaptionsepspace
    \else
      \ICRA@makecaption{#1}{#2}%
    \fi
  \endgroup
}
\makeatother

\title{\LARGE\bf DUET: Dual-Robot Understanding via Efficient Teaching}
\ifarxivversion
\author{
  Yiqi Zhao$^{1,\star}$,~Ruohai Ge$^{1,\star}$,~Celina Shiyu Wang$^1$,~Junjie Ye$^1$,~Muchen Xu$^1$,~Minhao Li$^1$,\\
  Sergey Zakharov$^2$,~Basile Van Hoorick$^2$,~Vitor Campagnolo Guizilini$^2$,~Leonidas Guibas$^3$,\\
  Gaurav S. Sukhatme$^1$,~Jyotirmoy V. Deshmukh$^1$,~Yue Wang$^1$\thanks{$\star$ Equal contribution. Order is determined by coin flip.} \thanks{YZ, RG, CSW, JY, MX, ML, GS, JVD, YW are with the Thomas Lord Department of Computer Science, University of Southern California. SZ, BVH, VCG are with Toyota Research Institute. LG is with Stanford University.}
}
\else
\author{Anonymous Authors}
\fi

\begin{document}
\ifarxivversion
\IEEEaftertitletext{\vspace{-4\baselineskip}}
\else
\IEEEaftertitletext{\vspace{-2\baselineskip}}
\fi

\maketitle
\input{sections/teaser}
\thispagestyle{empty}
\pagestyle{empty}
\bstctlcite{IEEEcompact}

\input{sections/abstract}
\input{sections/introduction}
\input{sections/related_works}
\input{sections/methods}
\input{sections/experimental_results}

\input{sections/limitations}
\input{sections/conclusion}

\bibliographystyle{bibtex/IEEEtran}
\bibliography{bibtex/bib}

\ifarxivversion
\input{sections/appendix.tex}
\fi

\end{document}

%% file: sections/teaser.tex
\begin{strip}
    \vspace{-8pt}
    \centering

    \includegraphics[width=\textwidth]{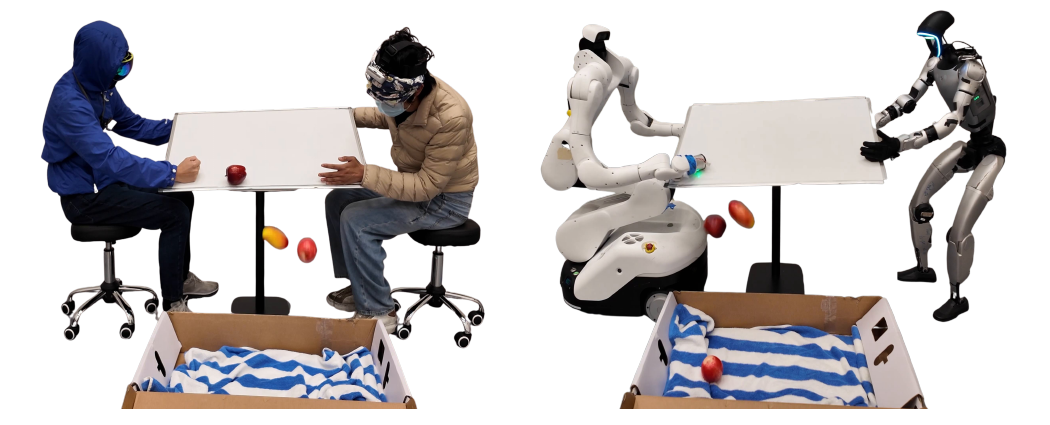}
    \par

    \ifarxivversion
        \vspace{3mm}
    \else
        \vspace{5mm}
    \fi

    \textbf{DUET.}
    We introduce a dual-robot policy learning framework that features
    efficient learning from human demonstrations.
    Left: Two humans perform a collaborative manipulation task.
    Right: A dual-robot system mimics the human teachers.
    \label{fig:teaser}
\end{strip}

%% file: sections/abstract.tex
\begin{abstract}
Dual-robot collaboration enables tasks that exceed the reach and payload of a single robot, such as collaboratively transporting objects across environments and executing coordinated handovers. Data acquisition is the primary bottleneck for training these systems. To this end, we introduce DUET, a dual-robot learning framework for mobile manipulation. For efficient data collection, we create a unified dual-embodiment synchronized VR-based teleoperation system for in-domain heterogeneous robot data collection. We further develop a complementary tracking pipeline that records human-human coordination and collaborative mobile manipulation priors. To allow efficient learning, we introduce an Action Chunking Transformer based architecture that first pretrains collaborative policies on efficient human-human demonstrations, before finetuning them on a minimal set of real-robot teleoperation trajectories. We develop a benchmark of four collaborative tasks to evaluate our framework using a Unitree G1 humanoid and a Dexmate Vega1 mobile manipulator. The results demonstrate that harnessing human priors not only yields superior task performance compared to baselines trained only on robot data, but also reduces the total human effort required for data collection. Our human data collection pipeline achieves $5.4 \times$ acceleration on average from teleoperation, but we perform better than robot-only data trained policies across all tasks. \ifarxivversion Our project page is available at \href{https://zhaoy37.github.io/Duet/}{https://zhaoy37.github.io/Duet/}.\fi
\end{abstract}
\ifarxivversion
\fi

%% file: sections/introduction.tex
\section{Introduction}
\begin{figure*}[t!]
    \centering
    \includegraphics[width=\textwidth]{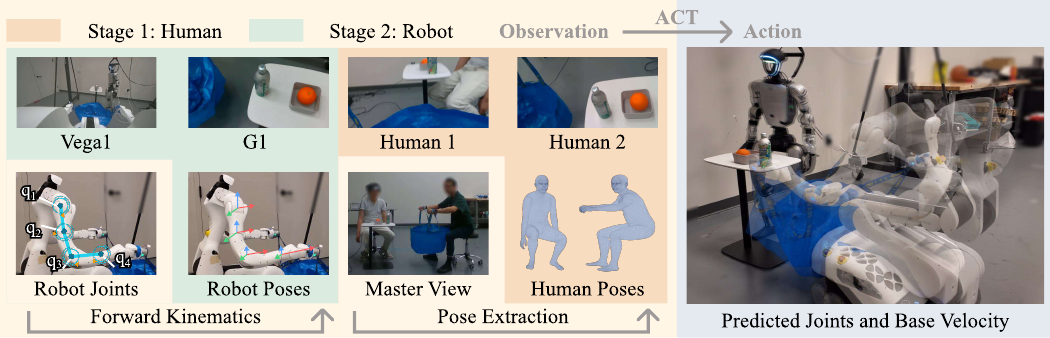}
    \caption{\textbf{DUET Overview.} In \textit{stage 1}, we pretrain an ACT architecture (Figure~\ref{fig:architecture}) with RGB streams from human-human demonstrations and extracted keypoints for proprioceptives. In \textit{stage 2}, in-distribution robot data finetunes the ACT policy.}
    \label{fig:training_pipeline}
\end{figure*}

\label{sec:introduction}
Many real-world tasks intrinsically exceed the kinematic reach and payload limits of a single robot \cite{fink2008multi}. Operations that require simultaneous, spatially separated interactions, such as handling oversized objects or executing asymmetric tool-use, make dual-robot systems a physical necessity rather than merely an algorithmic extension \cite{lai2025roboballet, wang2016multi}. Yet, realizing their practical utility requires advancing beyond collision-free navigation toward \textit{contact-rich collaborative mobile manipulation} \cite{fu2024mobile}. Controlling heterogeneous robot duos in this regime is not a trivial superposition of independent policies; it demands tightly coupled spatial-temporal coordination. For example, jointly manipulating cumbersome shared payloads demands strict dynamic equilibrium, while asymmetric interactions require tight temporal synchronization and shared perception. Because analytically modeling the multi-body dynamics of such joint operations is computationally prohibitive and brittle, end-to-end visuomotor policies \cite{kim2024openvla, ye2026world} have emerged as the compelling alternative, directly mapping high-dimensional observations to coordinated actions while bypassing explicit state estimation.

However, realizing the potential of these policies is severely bottlenecked by data acquisition. Orchestrating demonstrations for heterogeneous dual-robot systems typically demands complex setups and a large pool of human operators. To address this, we design a synchronized dual-operator teleoperation framework that allows just two human experts to successfully command the heterogeneous duo, a significant step forward in dual-robot data collection. Yet, even with this efficient system, teleoperation remains inherently limited by time and cost. To scale our dataset, we complement this framework with an efficient human-human data collection pipeline where two humans directly execute the collaborative tasks. Capturing direct human demonstrations \cite{kareer2025egomimic, hoque2026egodex, punamiya2026egoverse} is fundamentally faster and completely decouples data volume from robot hardware constraints, while naturally encoding rich, transferable priors regarding multi-robot spatial coordination. To synthesize these complementary streams, we introduce a pretraining paradigm that leverages low-cost human demonstrations to bootstrap the policy, requiring only a minimal set of high-fidelity dual-operator robot data. This approach effectively bridges the human-to-robot gap, maximizing the utility of both human agility and our streamlined dual-teleoperation system.

We introduce DUET (\textbf{D}ual-robot \textbf{U}nderstanding via \textbf{E}fficient \textbf{T}eaching), a unified centralized visuomotor policy framework for heterogeneous dual-robot collaboration  (Figure~\ref{fig:training_pipeline}). While supported by a dedicated dual-robot teleoperation pipeline, DUET fundamentally bypasses scaling limits by integrating efficient human demonstrations harnessing SAM 3D Body \cite{yang2026sam}. To bridge the resulting cross-embodiment gap, all data is projected into a shared pose space. Conditioned on these unified states and egocentric RGB streams, a single Action Chunking Transformer (ACT) \cite{zhao2023learning} backbone intrinsically enforces the tight spatial-temporal coupling demanded by contact-rich tasks. We validate DUET on a Unitree G1 humanoid (G1) and a Dexmate Vega1 mobile robot (Vega1) across a novel benchmark of four contact-rich tasks. Empirically, DUET demonstrates that integrating these human priors via collective training achieves better task performance compared to fully robot-demonstrated baselines, while requiring less overhead from teleoperation. \ifarxivversion \else Our code will be made publicly available. \fi We summarize our contributions:

\begin{enumerate}[label=\arabic*., leftmargin=*, topsep=4pt, itemsep=0pt, parsep=0pt]
    \item \textbf{Heterogeneous Dual-Operator Teleoperation:} We design a two-operator teleoperation framework to collect in-domain trajectories for a heterogeneous duo.
    \item \textbf{Human Collaboration Pipeline:} We propose a scalable data collection pipeline that captures direct human-human collaboration, ready for cross-embodiment pretraining.
    \item \textbf{Collective Training Architecture:} We harness an ACT-based framework that leverages human and robot data streams for collective training.
    \item \textbf{Benchmark and Evaluation:} We introduce a novel benchmark of four tasks including sweeping, transferring, balancing, and handovers. Empirically, pretraining on human demonstrations lowers overall data collection effort while achieving higher success rates than policies trained solely on larger robot datasets.
\end{enumerate}

%% file: sections/related_works.tex
\section{Related Work}
\label{sec:related_works}
\textbf{Imitation Learning and Teleoperation:} Imitation learning via human teleoperation has emerged as one of the premier paradigms for scaling physical interaction in open-world environments \cite{zhao2023learning, zhao2024aloha, khazatsky2024droid}. Recent breakthroughs have expanded this paradigm by integrating manipulation with mobile bases and legged locomotion, resulting in robust teleoperation pipelines for wheeled bimanual platforms \cite{fu2024mobile, dass2024telemoma, jiang2025brs}, scalable in-the-wild data collection interfaces \cite{chi2024umi, choi2026inthewild, etukuru2025rum}, and legged loco-manipulation frameworks for humanoids \cite{wei2026psi_0, luo2025sonic, jialong2025amo, li2025clone, ze2025twist2}. However, these infrastructures are designed for single-robot, leaving high-fidelity data collection for spatially distributed, heterogeneous multi-robot teams unresolved \cite{mattson2025r2bc, song2025collabot}. To bridge this gap, we introduce a portable, VR-based teleoperation framework for egocentric dual-robot data collection in mobile manipulation tasks.

\textbf{Multi-Robot Collaboration:} Multi-robot physical collaboration has a rich history, traditionally relying on classical control paradigms that demand complex state estimation, rigorous system modeling, and explicit communication architectures \cite{zhang2025multinonholonomic, michael2009cooperative, yang2022collaborative, fawcett2023distributed}. Seeking more adaptable behaviors, recent advances have increasingly shifted toward learning-based methods \cite{wang2022distributed, wu2024state}. Reinforcement learning can elicit sophisticated cooperative strategies in simulation, but typically depends on privileged, low-dimensional states and struggles to generalize to unstructured, visually rich real-world environments \cite{pandit2025multi, pandit2024learning, zeng2025decentralized, shibata2026dereco, chen2025chip}. Concurrently, end-to-end vision-action models have demonstrated remarkable generalizability for real-world manipulation tasks \cite{black2025pi, zhao2024aloha, zitkovich2023rt2}. To overcome the generalization limits of state-dependent control, we introduce a heterogeneous dual-robot framework directly mapping shared visual observations to synchronous joint actions.

\textbf{Learning from Human Demonstrations:} To alleviate the slow and costly nature of on-robot data collection, human demonstrations offer a scalable alternative \cite{punamiya2026egoverse}. Capturing rich physical affordances, human data provides a structural prior for policy learning. Recent frameworks leverage these priors via visual representation pre-training or cross-domain co-training to successfully transfer human dexterity to robots \cite{hoque2026egodex, kareer2025egomimic, heng2026humdex}. This paradigm has rapidly advanced complex embodiments, enabling view-invariant skill transfer \cite{fan2026robopaint}, in-the-wild humanoid loco-manipulation \cite{wei2026psi_0, shi2026egohumanoid, nai2026humanoid}, and interactive whole-body control via human demonstrations \cite{chen2026rhythm, huang2026learning, mao2025uh1}. However, while these priors have scaled single-robot learning, existing multi-robot extensions primarily target social or unconstrained kinematic behaviors. Utilizing synchronized multi-human data for physically coupled, object-centric manipulation remains unexplored. We introduce a multi-human pretraining pipeline, leveraging collaborative human data as a prior to accelerate and enhance our visuomotor policy.

%% file: sections/methods.tex
\section{Method}
\label{sec:methods}
\subsection{Teleoperation Pipeline}
\label{subsec:teleop}
\begin{figure}[t]
    \centering
    \includegraphics[width=\linewidth]{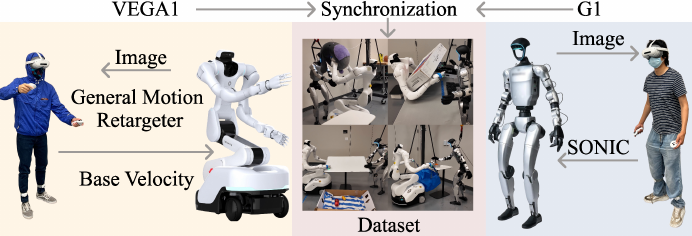}
    \caption{\textbf{Dual-Robot Teleoperation Pipeline.} Two human operators utilize PICO VR interfaces to simultaneously control a heterogeneous robot duo, Vega1 via General Motion Retargeting \cite{joao2025gmr} and G1 via the SONIC \cite{luo2025sonic} framework. The system streams real-time visual feedback while asynchronously logging joint pose data.}
    \label{fig:teleop_pipeline}
    \vspace{-3pt}
\end{figure}

As illustrated in Figure~\ref{fig:teleop_pipeline}, we introduce a synchronized teleoperation framework that enables just two human operators to simultaneously command a heterogeneous duo, demonstrated using a G1 and a Vega1. This system addresses the data acquisition bottleneck in multi-robot learning by providing a streamlined, low-latency interface for collaborative loco-manipulation tasks. By decoupling the human interface from the underlying execution layers, the architecture serves as a mature, hardware-agnostic foundation. It is capable of supporting any robotic embodiment with a compatible action retargeting module, allowing new platforms to be hot-swapped into the collaborative pipeline.

\textbf{Hardware and Visual Interface:} To ensure high-fidelity spatial awareness during physical interactions, operators utilize PICO VR headsets paired with handheld controllers and two ankle trackers. This setup serves as a comprehensive motion-tracking interface while simultaneously handling real-time egocentric video streaming via GStreamer. The captured human kinematic data is continuously streamed to the robot execution layers, while the visual feedback is explicitly tailored to each platform's sensory suite: the G1 operator receives a $1920 \times 1080$ (30 Hz) stream from a RealSense camera, while the Vega1 operator receives a wide-angle $2560 \times 720$ (30 Hz) stream from a ZED camera. To eliminate the standard requirement of a dedicated third operator to manage host PC interventions during rapid data collection, the VR controllers also serve as the central command hub, directly managing task initialization, environment resets, and data logging. The same streaming and human-human pose tracking pipeline supports any embodiment and camera source with compatible retargeting modules.

\textbf{Modular Retargeting:} Mapping strategies are tailored to the physical capabilities of each platform. For G1, we utilize the state-of-the-art SONIC \cite{luo2025sonic} framework for physics-based, whole-body motion retargeting. However, to adapt this for collaborative teleoperation, we introduce several critical extensions: we implemented the aforementioned real-time image streaming and integrated a 10 Hz asynchronous data collection module. Conversely, the control pipeline for Vega1 is custom-designed and relies on General Motion Retargeting (GMR) \cite{joao2025gmr} algorithm used for retargeting human motion to robot action, which can support other embodiments via customized configuration. We designed a GMR configuration that explicitly retargets upper-body human motion to Vega1 and mapped the PICO controller's analog stick to mobile base's $v_{x}, v_{y}$ and $\omega_{z}$. We implemented a lower-level inference controller utilizing Exponential Moving Average smoothing for stability, alongside visual and data logging modules similar to G1. Finally, to synchronize our collected dual-robot data, both robots log sensor and action states asynchronously at 10 Hz against host clocks synchronized via a local Chrony NTP server, enabling exact automated alignment during offline post-processing.
\begin{figure}[t]
    \centering
    \includegraphics[width=\linewidth]{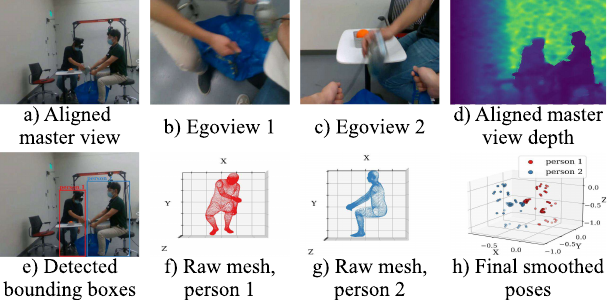}
    \caption{\textbf{Human Data Collection Example.} Our human-human demonstration pipeline processes multi-view RGB-D observations into calibrated 3D pose trajectories.}
    \label{fig:human_example}
    \vspace{-5pt}
\end{figure}
\begin{figure}[t]
    \centering
    \includegraphics[width=\linewidth]{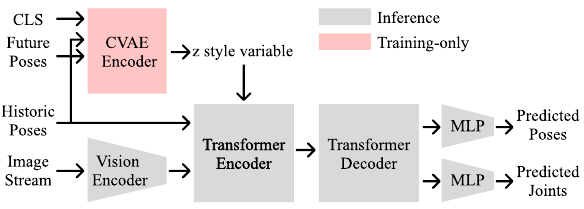}
    \caption{\textbf{Model Architecture.} Our ACT policy uses a $2$-head design with joint head randomly initialized during finetune.}
    \label{fig:architecture}
\end{figure}

\subsection{Human Data Collection Pipeline}
\label{subsec:human}
To complement the teleoperation system, we develop a pipeline that records two humans performing the same collaborative tasks without a robot in the loop. The recording setup consists of three cameras. We use one head/neck-mounted egoview camera for each operator and a fixed third-view \emph{master} camera positioned to keep both operators in frame, providing the input to subsequent offline human pose extraction following a soft cross-camera time alignment.

For each human clip, we start by extracting the poses of the two operators. We harness a YOLO-based \cite{redmon2016you} human detector to produce a set of $2$ bounding boxes tracking the human identities in the master frame. For each human $p$, SAM 3D Body \cite{yang2026sam} runs per frame for mesh extraction. It estimates anatomical keypoints $\mathbf{K}^{(p)}_t$, mesh vertices $\mathbf{V}^{(p)}_t$, and camera translation $\mathbf{c}^{(p)}_t$. Because these predictions lie in SAM's arbitrary-scale pseudo-camera frame, we first translate them into its predicted perspective frame,
\begin{align*}
    \widetilde{\mathbf{K}}^{(p)}_t = \mathbf{K}^{(p)}_t + \mathbf{c}^{(p)}_t,
  \widetilde{\mathbf{V}}^{(p)}_t = \mathbf{V}^{(p)}_t + \mathbf{c}^{(p)}_t.
\end{align*}
To lift the predictions into metric coordinates, we fuse the SAM predicted mesh $\widetilde{\mathbf{V}}^{(p)}_t$ with the master camera's depth stream. For each frame and tracked person, we render the mesh through SAM's internal focal length to obtain a predicted z-buffer $\hat{z}(u, v)$, restrict comparison to an eroded person silhouette, and recover the metric scale $s$ from the per-pixel ratio between the observed and predicted depth with a robust two-stage estimator 
\begin{align*}
    s_0 = \operatorname{median}_{(u,v) \in \mathcal{M}}\frac{z^{\mathrm{obs}}(u,v)}{\hat{z}(u,v)},  s = \frac{\sum_{(u,v) \in \mathcal{I}}
  z^{\mathrm{obs}}(u,v)\,\hat{z}(u,v)}
                {\sum_{(u,v) \in \mathcal{I}} \hat{z}(u,v)^2},
\end{align*}
\noindent where $\mathcal{M}$ is the eroded silhouette and $\mathcal{I} \subseteq \mathcal{M}$ contains pixels for which the rescaled prediction $s_0 \hat{z}$ agrees with the observation $z^{\mathrm{obs}}$ to within a tolerance threshold. We back-project SAM's 2D keypoints through the camera intrinsics at the recovered metric depths, yielding 3D poses in the physical camera frame of the master view. Lastly, a per-keypoint mask-weighted moving-average filter suppresses high-frequency jitter introduced by depth-sensor noise. We extract $9$ 3D keypoints (27 dimensions) per human and root velocities from pelvis movements, which we then use to pretrain our visuomotor policy.
\begin{figure*}[t!]
    \centering
    \includegraphics[width=\textwidth]{images/collection_time.pdf}
    \caption{%
\textbf{Benchmark data distribution and collection efficiency.}
\textbf{Left:} Total data span and number of episodes per pipeline across four tasks.
\textbf{Right:} \emph{Amortized collection time} (AT) for each task. The value above each task group is the \emph{speedup ratio}, defined as the teleoperation $AT$ divided by the human-demonstration $AT$, where $AT$ is the average time for each successful data collection.
}
\label{fig:collection_efficiency}
\end{figure*}
As shown in Figure \ref{fig:human_example}, this example shows the capacity of our human data collection pipeline in retargeting human data for collective training. We in sequence show example temporally aligned master and egoview camera images (together with the master view depth), example bounding boxes and meshes and example extracted poses. The poses in Figure \ref{fig:human_example} (h) are recentered so that the position of person 1's neck at the first valid frame of the clip is the origin.

\subsection{Collective Training}
\label{subsec:train}
Our robots are controllable via \textit{joint-space} commands. Our architecture (Figure~\ref{fig:architecture}) is inspired by \cite{kareer2025egomimic} and we train a single policy in two stages that share an ACT \cite{zhao2023learning} based architecture but differ in the data source. As depicted in Figure~\ref{fig:training_pipeline}, \textbf{Stage 1} pretrains the policy backbone on human-human collaboration data collected following Section~\ref{subsec:human}; \textbf{Stage 2} finetunes the resulting weights on robot teleoperation data of Section~\ref{subsec:teleop}. The two stages produce models with nearly identical parameter shapes, so all components except robot-specific projections are warm-started from human prior.

We consider an action representation where $[\hat{p}_{t + H}, \hat{q}_{t + H}]$ is the $H$ length predicted next-chunk position with $p_t \in \mathbb{R}^{d_p}$ encoding the \emph{(3D) pose} component shared by both embodiments and $q_t \in \mathbb{R}^{d_q}$ encoding the \emph{joint-space} component available only for robots. We use $\hat{p}$ to denote the prediction of ground-truth $p$ and similarly define $\hat{q}$. To bridge the embodiment gap, both modalities express $p_t$ in head-relative coordinates: Human poses $p_t^H$ recovered by Section~\ref{subsec:human} are re-centered at the operators' heads, while robot poses $p_t^R$ obtained from forward kinematics are expressed relative to each head link. For our heterogeneous duo, $d_p = 57$ comprises $27$ dimensions each of G1 and Vega1 keypoints and $3$ of Vega1 mobile base velocity command. On the human side, we derive the keypoints of the corresponding poses from Section~\ref{subsec:human} and estimate the base velocity from pelvis. The keypoint choice is a customizable design choice.

We utilize an ACT policy with a ResNet-18 \cite{he2016deep} visual encoder warm started from ImageNet \cite{russakovsky2015imagenet}, conditioned on a pair of historic egocentric RGB image streams (one per agent) and the corresponding pose-only proprioceptive states. To accommodate the heterogeneous action space between humans and robots, we adapt the architecture from \cite{kareer2025egomimic} and project output tokens of the ACT decoder backbone via two parallel multi-layer perceptrons that share the decoder's hidden state $h_t \in \mathbb{R}^{d_\text{model}}$ but predict separately $\hat{p}_{t+H} = \mathrm{MLP}_p(h_t)$ and $\hat{q}_{t + H} = \mathrm{MLP}_q(h_t)$. Both heads actively backpropagate on robot batches, while only the pose head is active during human batches. For structured regularization for the action-chunk decoder, our CVAE encoder ingests the pose-only subspace of the action chunk in both stages, ensuring consistent latent semantics during both stages. Adopting the standard \cite{zhao2023learning}, during inference, the CVAE encoder is bypassed and the decoder is conditioned on $z = \mathbf{0}$. The joint-space predicted pose and base velocity are forwarded for robot execution.

%% file: sections/experimental_results.tex
\section{Experimental Result}
\label{sec:results}
We evaluate the efficacy of DUET through a series of experiments designed to characterize both our data collection pipelines and the resulting visuomotor policies. Our empirical evaluation aims to answer the following research questions: \textbf{a) Q1:} How does the collection efficiency of direct human collaboration compare to our streamlined dual-robot teleoperation baseline? \textbf{b) Q2}: Does our dual-operator teleoperation pipeline serve as an effective source for high-fidelity dual-robot policy learning? \textbf{c) Q3:} Can our human data pretrained policy achieve better performance on mobile manipulation tasks while requiring less effort in data collection? We remark that the performance of our framework is dependent on our choice of the ACT architecture, substitutable with alternative visuomotor backbones without changing the data collection and pretraining framework. Our evaluation focuses on demonstrating the feasibility of learning end-to-end dual-robot visuomotor policies, while quantifying the gains achieved by integrating human priors.
\begin{figure*}[t!]
    \centering
    \includegraphics[width=\textwidth]{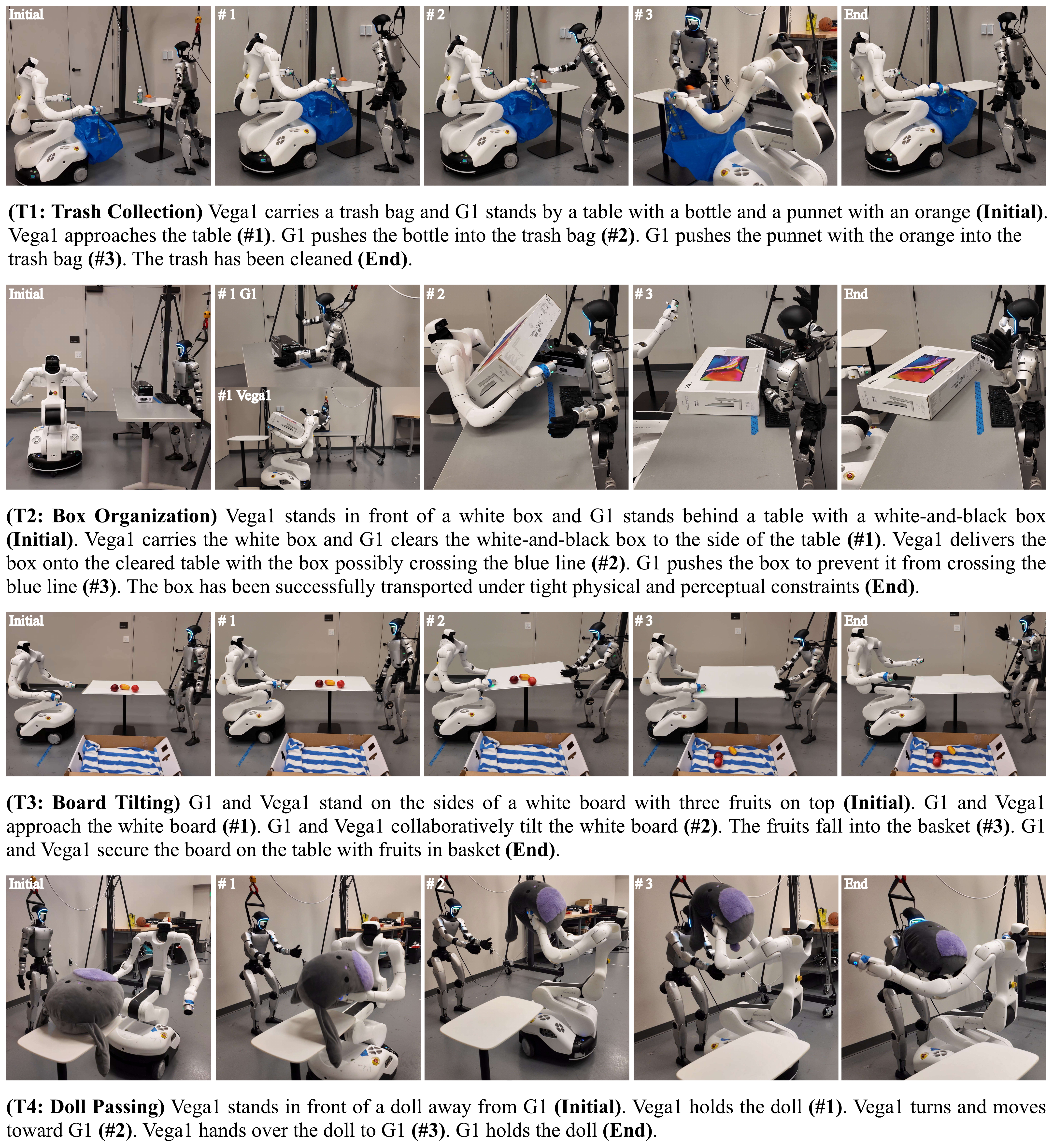}
    \caption{\textbf{Overview of the Tasks.} Snapshots show the initial setup and phases.}
    \label{fig:tasks}
    \vspace{-10pt}
\end{figure*}
\begin{figure}[t!]
    \centering

    \includegraphics[width=\linewidth]{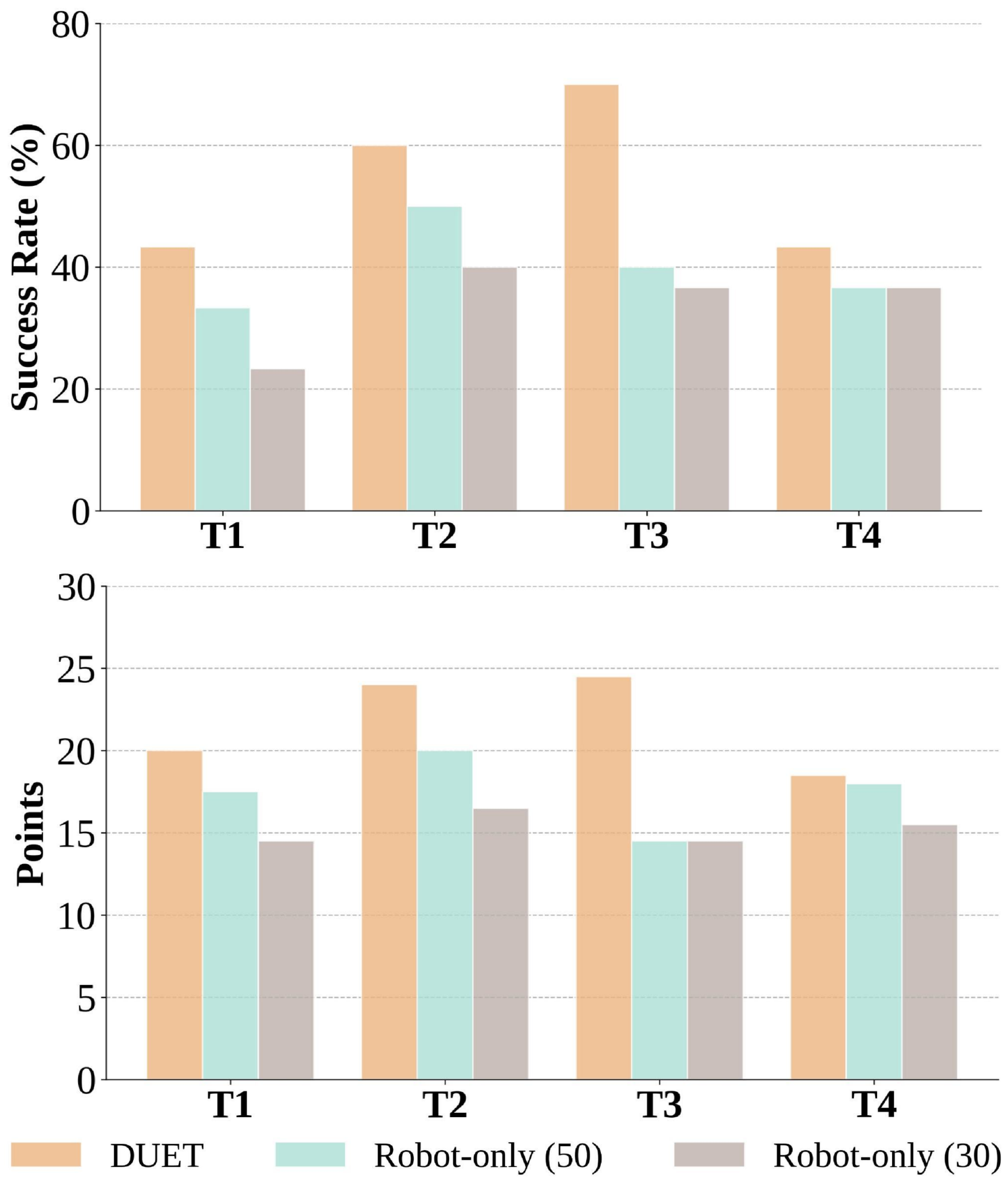}
    \caption{\textbf{Evaluation Results.} Comparison of task success rates and accumulated points over $30$ experiments for each task of T1–T4. DUET (60 human demos + 30 robot trajectories) is compared against robot-only baselines (trained on 30 and 50 trajectories).}
    \label{fig:results_composite}
    \vspace{-5pt}
\end{figure}
\subsection{Benchmark and Evaluation Setup}
\label{subsec:benchmark_eval_setup}
We introduce a benchmark suite of four tasks designed to isolate key multi-robot challenges, as detailed in Figure \ref{fig:tasks}. For each task, we use our teleoperation pipeline (Sec.~\ref{subsec:teleop}) to collect high-fidelity robot data on a heterogeneous platform, comprising Vega1 and G1, alongside corresponding human-human demonstrations (Sec.~\ref{subsec:human}): \textbf{a) T1: Trash Collection} focuses on \textit{asymmetric spatial-temporal coordination and time synchronization}. Vega1 navigates from a starting location to dynamically position a trash bag as G1 sweeps debris off a table. \textbf{b) T2: Box Organization} focuses on \textit{collaborative vision and long-horizon execution ($\sim$ 1 minute)}. G1 clears the workspace. Vega1 transfers a box to its perception boundary, where G1 then executes a corrective nudge. \textbf{c) T3: Board Tilting} focuses on \textit{collaborative manipulation and collective balance}. G1 and Vega1 cooperatively tilt a shared whiteboard to slide surface objects into a container, requiring precise force coordination to prevent the board from falling. \textbf{d) T4: Doll Passing} focuses on \textit{coordinated handovers and grasp transfer}. Vega1 secures and transfers a doll directly to G1, requiring synchronized release and capture maneuvers. The total number of collected episodes for each task across both teleoperation and human demonstration is detailed in Figure~\ref{fig:collection_efficiency}; \ifarxivversion The dataset will be released at \href{https://zhaoy37.github.io/Duet/}{https://zhaoy37.github.io/Duet/}.\else The dataset will be released.\fi{} To maintain a stable movement speed across the dataset, we account for the inherent kinematic overhead of teleoperation \cite{kareer2025egomimic}. Because human execution is naturally faster than the robots, the total duration of each human data clip is proportionally reduced by a factor of $1.5$ compared to the robot trajectories, and we proportionally scale the velocity in the training data.
\begin{table}[t]
    \centering
    \caption{\textbf{Data Collection Effort in Minutes.}}
    \label{tab:effort}
    \begin{tabularx}{\columnwidth}{l|*{4}{>{\centering\arraybackslash}X}}
        \toprule
        Method & \textbf{T1} & \textbf{T2} & \textbf{T3} & \textbf{T4} \\
        \midrule
        \rowcolor{blue!7}
        DUET (Ours)
            & $200.95$ & $206.50$
            & $111.65$ & $106.20$ \\
        Robot-only (50)
            & $268.42$ & $236.83$
            & $139.92$ & $116.00$ \\
        Robot-only (30)
            & $161.05$ & $142.10$
            & $83.95$ & $69.60$ \\
        \bottomrule
    \end{tabularx}
\end{table}
\subsection{Data Collection Efficiency Comparison}
\label{subsec:efficiency}
To address \textbf{Q1}, we evaluate data acquisition efficiency by comparing the \textit{Amortized Collection Time} ($AT$) of both pipelines, where $AT$ is the ratio between the total elapsed time devoted for data collection and the number of successfully acquired trajectories ($N = 10$) during this time. As shown in Figure~\ref{fig:collection_efficiency} (right), we report an $AT$ of \textbf{322.1}, \textbf{284.2}, \textbf{167.9}, and \textbf{139.2} seconds for teleoperation and an $AT$ of \textbf{39.9}, \textbf{64.4}, \textbf{27.7}, and \textbf{36.6} seconds for human demonstration respectively for \textbf{T1, T2, T3} and \textbf{T4}. While our teleoperation framework is highly optimized, with an average $AT$ of \textbf{228.4} seconds per task, direct human demonstration provides a \textbf{5.4}$\times$ acceleration on average. This underscores the advantages of human data collection: it improves collection efficiency while bypassing the high capital expenses of robot hardware and the severe cognitive bottlenecks inherent to teleoperation.
\begin{figure}[t!]
    \centering
    \includegraphics[width=\linewidth]{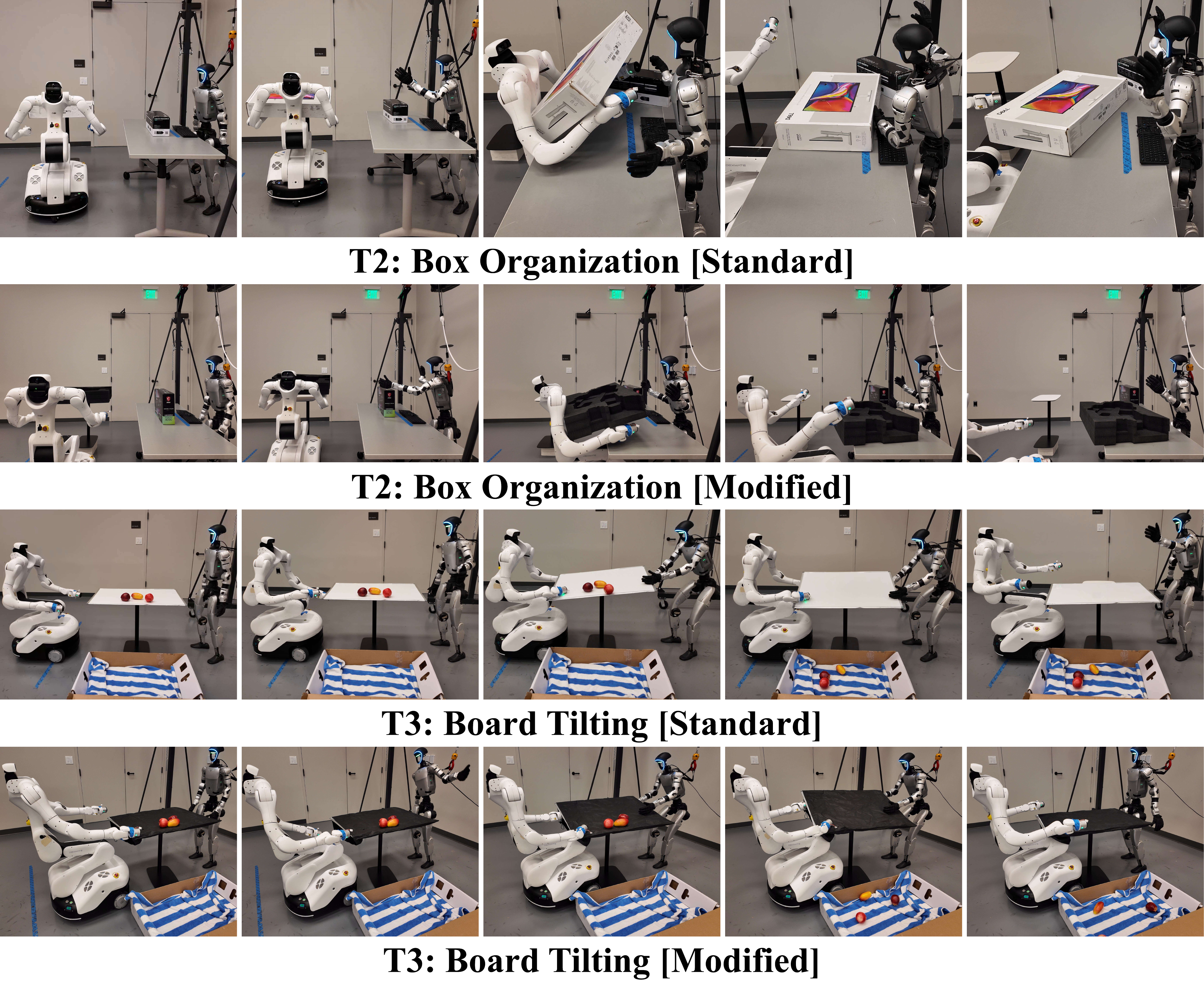}
    \caption{\textbf{Qualitative OOD Examples.} T2 changes box appearance and material; T3 covers the whiteboard in black.}
    \label{fig:generalizability}
    \vspace{-5pt}
\end{figure}
\subsection{Robot-only Ablation Study}
\label{subsec:robot_only_ablation}
To address \textbf{Q2}, we conduct an ablation study where only \textbf{Stage 2} from Section~\ref{subsec:train}, training with $50$ robot-only data for each task, is carried out on a fresh ACT (implemented through \cite{cadene2024lerobot}) without human prior.  To quantify multi-stage performance, we implement a normalized metric scoring each task trial from 0.0 to 1.0 based on sequential milestone completion. For \textbf{T1}, each piece of trash deposited into the basket yields 0.5 points for a total of two items. For \textbf{T2}, 0.5 points are awarded when G1 clears the workspace while Vega1 successfully secures the box, and the remaining 0.5 points are granted if the box is placed on the table without crossing the boundary line, which includes successful recoveries via a corrective nudge from G1. For \textbf{T3}, 0.5 points are awarded if the whiteboard remains on the table and at least one of the three items lands in the basket, with the full 1.0 point achieved only when all three items are safely contained. For \textbf{T4}, the initial object grasp by Vega1 and the subsequent dual-robot handover each contribute 0.5 points. To evaluate policy performance, we execute \textbf{30} independent physical hardware trials per task. As shown in Figure~\ref{fig:results_composite}, the policy yields overall success rates of \textbf{33.3\%, 50\%, 40\%, 36.7\%} respectively for \textbf{T1, T2, T3}, and \textbf{T4}. The cumulative scores across all 30 trials are \textbf{17.5, 20, 14.5}, and \textbf{18} respectively for the four tasks, demonstrating that our teleoperation pipeline yields a capable standalone baseline for multi-robot learning.

\subsection{DUET Performance}
To address \textbf{Q3}, we execute \textbf{Stage 1} (Section~\ref{subsec:train}) by pretraining on $60$ human demonstrations, followed by \textbf{Stage 2}, where we finetune using $30$ robot data. We evaluate these policies and present the resulting success rates and points in Figure~\ref{fig:results_composite}, where each experiment is conducted $30$ times. In this figure, we also compare our policy against the baseline policies trained on $30$ and $50$ robot only data without human prior. DUET and Robot-only (30) use the same $30$ robot trajectories; the Robot-only (50) training set is a strict superset. The results demonstrate a clear performance gain for the policy pretrained on human data across all evaluated tasks.

Drawing on the efficiency metrics established in Section~\ref{subsec:efficiency}, we compute the \textit{collection effort} ($E$) for each policy, measuring the total expected time required to acquire its underlying training dataset. We report these metrics in Table~\ref{tab:effort}. Formally, effort is defined as $E = \frac{N_H \cdot AT_H + N_R \cdot AT_R}{60}$, where $N_H$ and $N_R$ denote the number of human and robot training trajectories used, while $AT_H$ and $AT_R$ represent their respective amortized collection times in seconds computed in Section~\ref{subsec:efficiency}. The factor of $60$ converts the total collection effort to minutes. We show that while DUET achieves better performance, its training requires less collection effort than policies trained on $50$ robot-only data. Qualitatively, the human-pretrained policy exhibits smoother execution and better adheres to object affordance priors implicitly captured during human data collection. These factors largely account for its higher success rate compared to the baseline.

\input{sections/generalization}

%% file: sections/generalization.tex
\subsection{Qualitative OOD Case Study}
\label{subsec:qualitative_generalization}
To provide a qualitative indication of behavior under distribution shift, we conduct OOD case studies on  \textbf{T2 : Box Organization} and \textbf{T3 : Board Tilting}. These experiments are intended as illustrative examples rather than a quantitative evaluation of generalization. For T2, we introduced perturbations by replacing the standard white box with a black foam box and altering the G1 robot's manipulation white-and-black box to a green-and-black box. In T3, we obscured the white board with a black cover to shift the visual distribution. As illustrated in Figure \ref{fig:generalizability}, these examples suggest that the learned policy is not strictly tied to training-time object appearance.

%% file: sections/limitations.tex
\section{Limitation}
\label{sec:limitations}
While our framework is model-agnostic, we only experimented on the ACT architecture. Further explorations with other state-of-the-art architectures, which may improve the pretraining, are left as future work. Our evaluation is restricted to one heterogeneous pair. Our pipeline is designed to be hardware-agnostic, and extending DUET to a wider range of embodiments and larger robot teams is a natural next step toward general-purpose, scalable multi-robot collaboration.

%% file: sections/conclusion.tex
\section{Conclusion}
\label{sec:conclusion}
In this work, we introduced a multi-human pre-training pipeline, leveraging collaborative human data as a prior to accelerate and enhance our dual-robot visuomotor policy. Our evaluations demonstrate that policies pre-trained on human data yield task performance that exceeds baselines trained exclusively on robot data of comparable collection time. Furthermore, we observe that incorporating human data can yield high motion smoothness, attributable to the diverse and smooth nature of human movement. A primary takeaway is that, despite the inherent domain gap, human demonstrations can effectively reduce our reliance on robot data through an appropriate pre-training framework. More broadly, this underscores the immense potential of large-scale human data as a pre-training foundation for multi-robot design. Building on this paradigm, future systems can seamlessly adapt these multi-human priors to diverse embodiments via minimal fine-tuning, paving the way for scalable coordination across any number of robots.

\ifarxivversion
\section{Acknowledgments}
\label{sec:acknowledgments}

We thank Cameron Smith for valuable discussions on 3D computer vision and general robot learning. We thank Hongyi Jing and Yuzhe Qin for sharing their expertise and answering our questions about the robots, and Enze Li and Kai Liang for their help with robot repairs. We are also grateful to everyone in the RESL Lab at USC and Yihe Tang for their valuable discussions and feedback on paper writing. The USC Physical Superintelligence Lab acknowledges generous support from Toyota Research Institute, Dolby, Google DeepMind, Capital One, Nvidia, Bosch, NSF, and Qualcomm. This work was partially supported by the National Science Foundation through the following grants: CAREER award (SHF-2048094), IIS-SLES-2417075, and funding by Toyota R\&D through the USC Center for Autonomy and AI. Yue Wang is supported by a Powell Research Award.
\else
\fi

%% file: sections/appendix.tex
\tableofcontents
\begin{table*}[t]
    \centering
    \caption{\textbf{Controller Mapping.} Comprehensive input mapping detailing how operator commands are translated into specific robotic behaviors and system operations for both the G1 and Vega1 robots.}
    \begin{tabular}{lll}
        \toprule
        \textbf{Task} & \textbf{G1 Controller Mapping} & \textbf{Vega1 Controller Mapping} \\
        \midrule
        Toggle Teleoperation (Start/Pause) & Menu & X \\
        \midrule
        Record Data & Right Grip + A & Y \\
        \midrule
        Reset to Initial Pose and Pause & A + X & A \\
        \midrule
        Camera View (Zoom in/Out) & B & B \\
        \midrule
        Trigger & Close Hand & Close Hand \\
        \midrule
        Grip & Open Hand & Open Hand \\
        \midrule
        Keep Recorded Data & Left Grip + X & Left Axis Click \\
        \midrule
        Yaw Control ($\omega_{z}$) & N/A & Left Axis \\
        \midrule
        Velocity Control ($v_{x}, v_{y}$) & N/A & Right Axis \\
        \bottomrule
    \end{tabular}
    \label{tab:pico_mapping}
\end{table*}
\section{Dual Robot Teleoperation System}

\subsection{Extended Teleoperation Implementation Details}
\label{subsec:teleoperation_system_details}
Building upon the dual-robot teleoperation interface introduced in Section \ref{subsec:teleop}, this section provides additional technical details regarding system-wide control and transmission frequencies, lower-level smoothing algorithms, and hand-retargeting mechanics. The teleoperation interface relies on a pair of PICO4 Ultra VRs, which capture operators' motions as full-body human pose estimates in SMPL format using the headsets, handheld controllers, and two pairs of ankle trackers. The raw full-body human poses are retargeted and continuously transmitted to both robots at 50 Hz. Once processed, these retargeted joint-space commands are executed with a control frequency of 500 Hz.

Because raw human movement contains high-frequency jitter that can destabilize physical hardware, each robot's lower-level controllers process this stream differently. For G1, the SONIC \cite{luo2025sonic} framework manages these high-frequency commands to maintain whole-body balance and stability. Conversely, for the Vega1 mobile manipulator, we engineered a custom lower-level inference controller that applies Exponential Moving Average (EMA) smoothing to the command stream using the update rule $S_t = \alpha x_t + (1 - \alpha) S_{t-1}$, where $S_t$ and $S_{t-1}$ represent the current and previous smoothed actions, $x_t$ is the raw incoming command, and the smoothing factor $\alpha$ is set to $0.01$.

While G1's hand actuation is natively managed within SONIC, we designed a simplified hand retargeting approach for the Vega1 mobile manipulator. Similar to TWIST2 \cite{ze2025twist2}, instead of continuous hand pose estimation, we treat the F5D6 hand of Vega1 entirely as a gripper. By decoupling complex finger tracking from the interface, operators control the hand's actuation cleanly and reliably via button presses on the PICO handheld controllers. The VR controllers capture the operator's gripping intent as a continuous signal, which the host software translates into a normalized scalar grasp command, $\gamma \in [0, 1]$. In this mapping, $\gamma = 0$ denotes a fully open hand and $\gamma = 1$ represents a closed hand. The controller relies on two predefined joint configurations: an open pose ($q_{\rm open}$) and a close pose ($q_{\rm close}$). The final commanded hand joint configuration is continuously computed via linear interpolation: $q_{\rm hand} = (1 - \gamma) q_{\rm open} + \gamma q_{\rm close}$.

\subsection{Dual-Operator Interface Mapping}

To empower two operators to fully manage the heterogeneous dual-robot teleoperation system independently, eliminating the need for a dedicated host PC supervisor, we designed a mapping of the controller inputs to all essential robotic and system-level actions (detailed in Table \ref{tab:pico_mapping}).

\subsection{Rationale for Retargetting Frameworks}

The General Motion Retargeting (GMR) \cite{joao2025gmr} algorithm provides a highly effective, universal mapping solution for cross-embodiment control. Because the Vega1 mobile manipulator requires only upper-body articulation on a stable wheeled base, we purpose-built its control pipeline to leverage GMR. However, direct kinematic retargeting is insufficient for the G1 humanoid. Retargeting human motion to the G1 humanoid requires an extra layer of control to ensure lower-body stability and high-fidelity physical resemblance, which we achieve by utilizing SONIC \cite{luo2025sonic}, the state-of-the-art framework for physics-based, whole-body teleoperation for the G1 humanoid.

\subsection{Robot Data Collection}
To facilitate temporal alignment across domains, clip durations are fixed on a per-task, per-domain basis. Specifically, recording durations are fixed on a strictly per-task basis. For any given task, all robot teleoperation data clips share a uniform duration, while all human-human demonstrations share a standardized duration that is proportionally reduced by a factor of 1.5, as introduced in Section \ref{subsec:benchmark_eval_setup}. Clip length for each task are detailed in Table \ref{tab:robot_parameters}. This approach is motivated by the need to enforce synchronized motion representations \cite{kareer2025egomimic}, which empirically improves velocity predictions and preserves motion smoothness when incorporating human pretraining priors. To align the dual robot data streams offline, we rely on common timestamps synchronized via a centralized host computer running a Chrony NTP server. This post-alignment procedure allows us to accurately synchronize the data from both robots, resulting in a cohesive 10 Hz dataset comprising time-aligned egocentric images, joint-space poses, and Vega1 base velocities.

\subsection{Teleoperation System Example}

\begin{figure*}[t!]
    \centering
    \includegraphics[width=\textwidth]{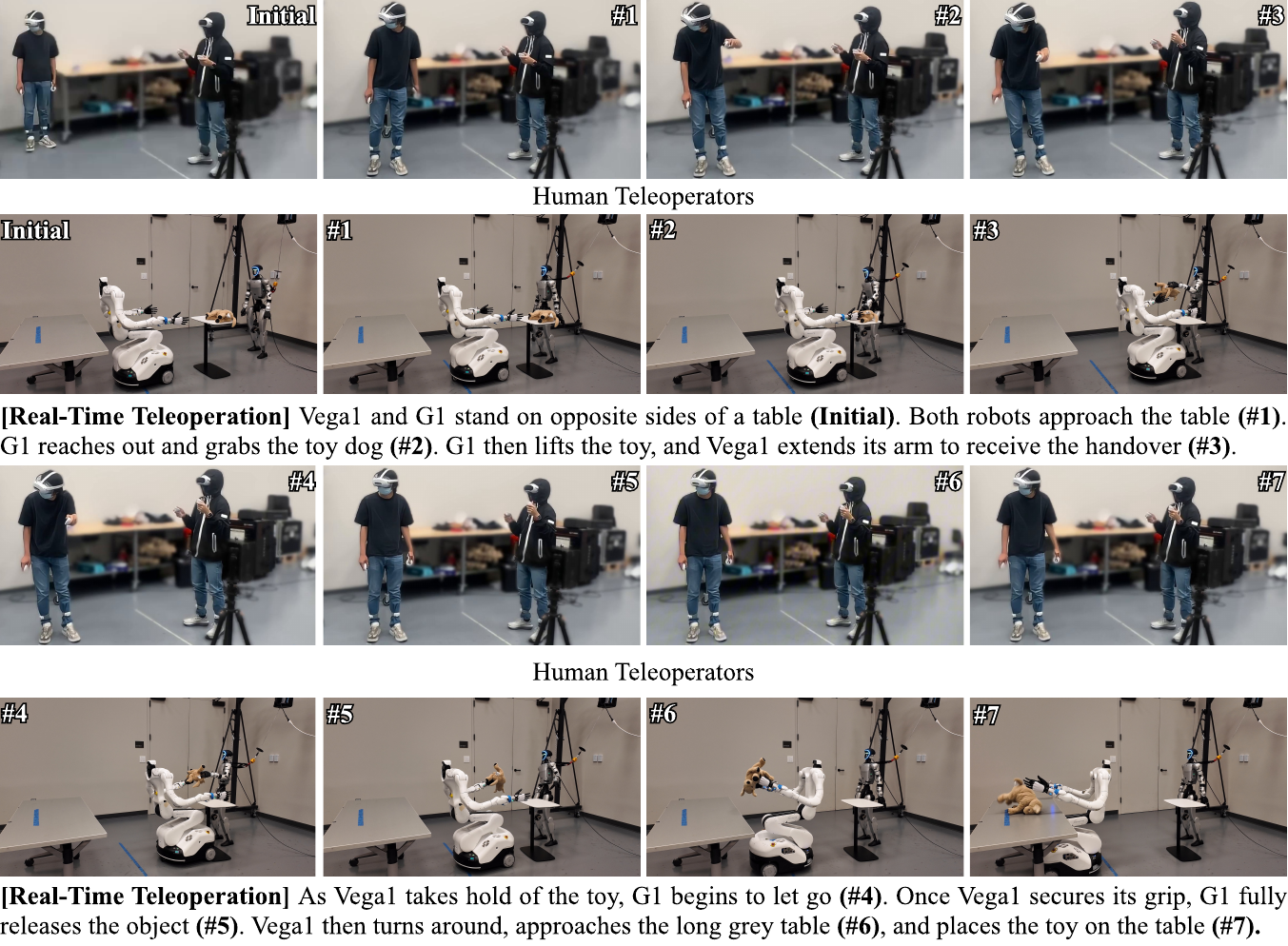}
    \caption{\textbf{Teleoperation System Example.} A side-by-side of the human teleoperator and the synchronized dual-robot execution. The sequence demonstrates the G1 humanoid walking to the workspace to grasp the target object (a toy dog), followed by a coordinated physical handover to the Vega1 mobile manipulator, which subsequently pivots and places the object onto a secondary table.}
    \label{fig:teleoperation_example}
\end{figure*}

As shown in Figure \ref{fig:teleoperation_example}, this example demonstrates our teleoperation system's ability to successfully capture complex human tasks and translate them directly to the robots in real time. The figure illustrates the synchronized execution of a collaborative loco-manipulation task alongside the operators.

To provide further details regarding the amortized collection times shown in Figure \ref{fig:collection_efficiency}, we outline the specific operational protocols used for each setup. To optimize the robotic data gathering process, tasks were executed by our most experienced teleoperator, who was occasionally supported by a third-party assistant dedicated to rapidly resetting the workspace between trials. The human-human demonstration data was collected by an arbitrary demonstrator operating entirely independently, managing all tasks and environment resets without external assistance.

\section{Human Data Collection Pipeline}
\label{sec:human_detail}

In this section, we detail the human data collection with hyperparameter choices listed in Table ~\ref{tab:human_parameters}.

\subsection{Video capture and alignment}
\label{subsec:video_capture}

Multi-view RGB-Depth capture uses an array of Intel RealSense D-series cameras, where one camera is the master and the rest two placed on head/neck mounts of the humans. The master camera captures the two humans in its frame during the task. Each camera streams color and depth at $W \times H = 640\times 480$ at $15$ Hz. Each clip is consistently $T$ seconds long for each task. During recording, the depth frame passes through a spatial, a temporal, and a hole filling filter. For each camera $c_i$ (where $c_m$ denotes the master camera and $c_1, c_2$ denote the egoview cameras), we also capture the camera intrinsics including focal lengths $(f_x^{c_i}, f_y^{c_i})$ and the principal points $(c_x^{c_i}, c_y^{c_i})$. After recording, we perform a nearest-neighbor alignment between the cameras.

\subsection{Mesh Extraction} 

\label{subsec:mesh}
From each clip's aligned master-camera RGB streams, we offline extract per-frame, per-person 3D anatomical skeletons and body meshes through a two-stage detect-then-fit pipeline. In \textbf{the detection and tracking stage}, every color frame is passed through a YOLOv8-nano \cite{redmon2016you} detector restricted to the COCO person class. The detections are streamed through the ByteTrack multi-object tracker to track identifiers across frames. A semantic identity assignment is performed once and frozen for the remainder of the clip: the active tracks are sorted in ascending order of bounding-box center $x$ coordinate, allowing training-time correspondence binding each role (G1, Vega1) to a fixed demonstrator. For invalid detections, we reuse bounding boxes from previous frames optionally. In the \textbf{mesh fitting stage}, the bounding boxes are fed into the SAM 3D Body DINOv3 \cite{yang2026sam} estimator, which returns per person and per frame, (i) the 3D anatomical keypoints $\mathbf{K}^{(p)}_t \in \mathbb{R}^{70 \times 3}$ in the mhr70 convention (ii)  the dense body-mesh vertices $\mathbf{V}^{(p)}_t \in \mathbb{R}^{N_V \times 3}$; (iii) the per-keypoint 2D image-plane projection $\mathbf{u}^{(p)}_t \in \mathbb{R}^{70 \times 2}$ used by the calibration stage, (iv) the predicted camera translation $\mathbf{c}^{(p)}_t\in \mathbb{R}^{3}$ that places the SMPL-style canonical body inside a perspective camera; and (v) the predicted focal length $f^{(p)}_t \in \mathbb{R}^+$ of that perspective model. The raw-pose tensors live in the pseudo-camera frame of the SAM 3D Body model rather than the master camera frame, and thus we carry out a depth-based calibration step.

\subsection{Depth-based Calibration}
\label{subsec:calib}
For calibration, every frame $t$ and every semantic identity $p$ is processed as follows. We first lift the SAM keypoints and mesh vertices out of the model's pose only canonical frame into its perspective camera-frame by adding the predicted camera translation
\begin{align*}
    \widetilde{\mathbf{K}}^{(p)}_t = \mathbf{K}^{(p)}_t + \mathbf{c}^{(p)}_t,
  \widetilde{\mathbf{V}}^{(p)}_t = \mathbf{V}^{(p)}_t + \mathbf{c}^{(p)}_t.
\end{align*}
 The lifted mesh vertices are then projected through the SAM perspective intrinsics $(f^{(p)}_t, f^{(p)}_t, W/2, H/2)$ onto the image plane and aggregated into a per-pixel 
  z-buffer retaining the minimum $z$ at each pixel as the implicit depth of the body surface viewed from the SAM perspective camera viewpoint: Concretely, for each vertex $\mathbf{v}_k = (x_k, y_k, z_k) \in \widetilde{\mathbf{V}}^{(p)}_t$ the per-pixel projection is
  $$
  u_k = \operatorname{round}\left(f^{(p)}_t \cdot \tfrac{x_k}{z_k + \varepsilon} + \tfrac{W}{2}\right),
  \qquad
  v_k = \operatorname{round}\left(f^{(p)}_t \cdot \tfrac{y_k}{z_k + \varepsilon} + \tfrac{H}{2}\right),
  $$
  with $\varepsilon = 10^{-9}$, and the z-buffer aggregates only the closest vertex per pixel, $z_\text{sam}(u, v) = \min\{z_k : (u_k, v_k) = (u, v), z_k > 10^{-6}, 0 \le u_k < W, 0 \le v_k < H ,\},$
 with $z_\text{sam}(u, v) = +\infty$ at pixels onto which no in-bounds vertex projects. The observed metric depth, $z_\text{obs}(u, v)$, from the master RealSense sensor is recovered from the captured depth image. The two depths are compared inside an eroded person mask $\mathcal{M}^{(p)}_t$, the per-frame bounding-box rectangle shrunk by a $5{\times}5$ kernel for one iteration to discard boundary pixels likely to straddle the body silhouette and the background. Letting
  $\mathcal{P}^{(p)}_t$ denote the pixels satisfying the joint validity criterion 
  $(u, v) \in \mathcal{M}^{(p)}_t \wedge 10^{-6} < z_\text{sam}(u, v) < +\infty \wedge \text{depth image}(u, v) > 0,
  $ an initial scale factor is recovered as the element-wise median of observed-to-rendered depth ratios,
  $$
  s_0^{(p)}[t] = \operatorname{median}_{(u, v) \in \mathcal{P}^{(p)}_t} \frac{z_\text{obs}(u, v)}{\max\big(z_\text{sam}(u, v),
  10^{-8}\big)}.
  $$ 
  The set $\mathcal{P}^{(p)}_t$ is required to contain at least $500$ pixels; if the eroded mask is too aggressive, the criterion is relaxed by replacing $\mathcal{M}^{(p)}_t$ with the un-eroded rectangle. Given a valid initial estimate, we sharpen it with a least-squares refit on the inlier subset defined by twin tolerance bands around the
  predicted depth $\hat{z}_\text{sam}(u, v) = s_0^{(p)}[t] \cdot z_\text{sam}(u, v)$, where
  $
  \mathcal{I}^{(p)}_t = \{(u, v) \in \mathcal{P}^{(p)}_t : z_\text{obs}(u, v) \ge \hat{z}_\text{sam}(u, v) - \tau_\text{front} \wedge
  |z_\text{obs}(u, v) - \hat{z}_\text{sam}(u, v)| < \tau_\text{abs}\},
  $ with $\tau_\text{front} = 0.10\text{m}$, allowing the observed depth to lie up to $10$ cm in front of the rendered surface and thereby
  absorbing thin self-occluders, and $\tau_\text{abs} = 0.30\text{m}$, rejecting any pixel whose depth disagrees by more than $30$ cm.
  Provided $|\mathcal{I}^{(p)}_t| \ge 20$, the refined scale is the closed-form linear-least-squares ratio
   $$
  s^{(p)}[t] = \frac{\sum_{(u, v) \in \mathcal{I}^{(p)}_t} z_\text{obs}(u, v) \cdot z_\text{sam}(u, v)}{\max\Big(\sum_{(u, v) \in
  \mathcal{I}^{(p)}_t} z_\text{sam}(u, v)^2, 10^{-12}\Big)},
  $$
 which replaces $s_0^{(p)}[t]$; otherwise the initial median estimate is kept unchanged. Letting $\widetilde{z}^{(p)}_{t,j}$ denote the $z$-component of the lifted keypoint $\widetilde{\mathbf{K}}^{(p)}_{t,j}$, the recovered scale converts it into RealSense-metric units,
 $$
  z^{(p)}_{t,j} = s^{(p)}[t] \cdot \widetilde{z}^{(p)}_{t,j}.
  $$
We then anchor each keypoint into the master color camera's intrinsic frame by treating the SAM-predicted 2D image-plane projection $\mathbf{u}^{(p)}_t = {(u_{t,j}, v_{t,j})}_{j=1}^{70}$ as observed pixel coordinates in the RealSense color image and $z^{(p)}_{t,j}$ as the corresponding metric depth, and back-projecting through the master color pinhole as
$$
  \mathbf{p}^{(p)}_{t,j} = \Bigg(\frac{(u_{t,j} - c^{c_m}_x)z^{(p)}_{t,j}}{f^{c_m}_x},; \frac{(v_{t,j} - c^{c_m}_y)z^{(p)}_{t,j}}{f^{c_m}_y}, z^{(p)}_{t,j}\Bigg).
  $$
A final axis re-labeling rotates the result from the RealSense optical convention $(X_\text{right}, Y_\text{down}, Z_\text{forward})$ into the visualization/world convention $(X_\text{forward}, Y_\text{left}, Z_\text{up})$ used everywhere downstream. We use the same notation $\mathbf{p}^{(p)}_{t,j}$ to denote the final depth calibrated poses without loss of generality. The calibrated keypoint trajectories contain jitter from depth noises. We address this with a moving average applied independently per person, per joint, and per coordinate, awaring the validity of the extracted keypoints.

\section{Data Preprocessing and Model Architecture}
In this section, we introduce our model architecture, along with how we process the raw data. See our choice of hyperparameters in Table \ref{tab:training_parameters}.
\begin{table*}[h!]
    \centering
    \caption{\textbf{Real-World Benchmarking Results.} Detailed evaluation reporting partial sub-task progress and overall success rates and points. A trial is considered successful only when all constituent sub-tasks are completed. Blue highlights denote the DUET performance in each category.}
    \resizebox{0.8\textwidth}{!}{
    \begin{tabular}{l|ccccc}
        \toprule
        \textbf{T1 : Trash Collection} & Single Item & Both Items & Points $\uparrow$ & Success Rate $\uparrow$\\
        \midrule
        \rowcolor{blue!7}
        DUET (Ours) & 27 & 13 & 20 & 13/30 \\
        Robot-only (50) & 25 & 10 & 17.5 & 10/30 \\
        Robot-only (30) & 22 & 7 & 14.5 & 7/30 \\
        \toprule
        \textbf{T2: Box Organization} & Clear \& Secure & In-Bound Placement & Points $\uparrow$ & Success Rate $\uparrow$ \\
        \midrule
        \rowcolor{blue!7}
        DUET (Ours) & 30 & 18 & 24 & 18/30 \\
        Robot-only (50) & 25 & 15 & 20 & 15/30 \\
        Robot-only (30) & 21 & 12 & 16.5 & 12/30 \\
        \toprule
        \textbf{T3 : Board Tilting} & Partial Containment & Full Containment & Points $\uparrow$ & Success Rate $\uparrow$ \\
        \midrule
        \rowcolor{blue!7}
        DUET (Ours) & 28 & 21 & 24.5 & 21/30 \\
        Robot-only (50) & 17 & 12 & 14.5 & 12/30 \\
        Robot-only (30) & 18 & 11 & 14.5 & 11/30 \\
        \bottomrule
        \textbf{T4: Doll Passing} & Initial Grasp & Successful Handover & Points $\uparrow$ & Success Rate $\uparrow$ \\
        \midrule
        \rowcolor{blue!7}
        DUET (Ours) & 24 & 13 & 18.5 & 13/30 \\
        Robot-only (50) & 25 & 11 & 18 & 11/30 \\
        Robot-only (30) & 20 & 11 & 15.5 & 11/30 \\
        \bottomrule
    \end{tabular}
    }
    \label{tab:detailed_metric}
\end{table*}

\subsection{Data Preprocessing}
\label{subsec:data_preprocessing}
We introduce our data preprocessing.  During training, our model takes in a $K$-step history of observations. On the robot side, we resize the historic image stream into size $(K, 2, 224, 224, 3)$ where $2$ robots each input $K$ steps of egoview RGB images of size $224 \times 224 \times 3$. In addition to the image stream, the policy is conditioned on a $K$-step history of poses, generated via forward kinematics using each robot's URDF and recorded joint states. We recenter the poses in root-relative coordinates by subtracting the position of each robot's head joint. From the full skeleton, we retain a fixed subset of $9$ joints per robot. For G1, we use the keypoints in sequence of head, left wrist pitch, right wrist pitch, left elbow, right elbow, left shoulder pitch, right shoulder pitch, left hip pitch, and right hip pitch. For Vega1, we use the keypoints in sequence of head joint $3$, left arm joint $7$, right arm joint $7$, left arm joint $4$, right arm joint $4$, left arm joint $1$, right arm joint $1$, and we repeat the entries of torso joint $3$ twice to match the input dimensions with the human input. For Vega1, we additionally include a velocity pseudo-joint that encodes the mobile base's planar velocity command $(v_x, v_y, \omega_z)$ as a $3$-vector. Together, the $K$-step history of poses is of shape $(K, 57)$ where each keypoint pose is a $3$-dimensional vector. We flatten this high-dimensional pose vector to $K \times 57$ before being projected into the transformer's hidden dimension. The ground truth output of the model is a chunk of length $H$ consisting of a future pose and a future joint-space (qpose) block. The pose block mirrors the structure of the state stream exactly. The qpose block consists of $23$ joint angles from G1 and $18$ from Vega1. Unlike the pose block, qpose is never observed as part of the policy's input; it appears only as a prediction target on the output side, providing low-level command signal that ultimately drives the physical actuators. Together, the shape of the prediction target is $H \times 98$.

On the human side, demonstrations are provided as per-clip 3D keypoint trajectories on the mhr70 skeleton expressed in a shared world frame (See Section~\ref{sec:human_detail}). Each human represents a robot. The two egoview RGB streams associated with assigned camera serials are also resized to $224 \times 224$ to match the robot side. For each of the two designated persons, every keypoint position has the neck position subtracted, in direct analogy to the robot side. The camera's frame is not naturally aligned with the kinematics frame that the robots induce, so for each human we apply a single fixed rotation matrix $R_\text{role} \in SO(3)$, parameterized by intrinsic XYZ Euler angles in degrees and applied to every keypoint at every timestep. Each $R_\text{role}$ is chosen by inspecting the subject's orientation in the first frame of a representative clip and is held constant across all frames. It is not re-estimated per frame and therefore does not track the subject's rotation across each clip. This optional design choice, however, empirically shows that human priors are largely rotation-invariant. From the $70$ mhr70 keypoints, we retain the same nine anatomical joints for both persons: neck, left and right wrists, left and right elbows, left and right shoulders, and left and right hips. For the person representing Vega1, we additionally compute a velocity pseudo-joint $(v_x, v_y, w_z)$ from finite differences of the rotated but uncentered pelvis position (taken as the midpoint of the left and right hip keypoints), where $\omega_z$ is the finite difference of the inter-hip vector's azimuthal angle. The resulting vector is multiplied by a scalar of $0.667$ so that the magnitude matches the dexmate odometry to account for the differences in human vs. robot speed during data collection. In summary, the $K$-step pose history is of shape $(K, 57)$ (flattened for projection into the transformer's hidden dimension). The accompanying $K$-step image history is of shape $(K, 2, 224, 224, 3)$. Unlike the robot side, the ground truth prediction target is a chunk of length $H$ consisting of only the pose block of size $H \times 57$. The qpose block is absent because human demonstrations do not carry joint-angle measurements.

All inputs and targets are standardized before being projected into the transformer. Following \cite{kareer2025egomimic}, we compute separate normalization statistics for the robot and human data. For the image stream, each frame is first rescaled from the discrete pixel range to $[0, 1]$, then standardized per channel using the ImageNet \cite{russakovsky2015imagenet} statistics $\boldsymbol{\mu}\text{img} = (0.485, 0.456, 0.406)$ and $\boldsymbol{\sigma}\text{img} = (0.229, 0.224, 0.225)$, applied as $\mathbf{x} \mapsto (\mathbf{x} - \boldsymbol{\mu}\text{img})/\boldsymbol{\sigma}\text{img}$ over the channel axis. For the state stream and the action target, we apply per-dimension $z$-score standardization with mean and standard deviation estimated empirically on the training split of each modality.

\subsection{Architecture Detail}

The main architecture is implemented with \cite{cadene2024lerobot} following a standard ACT procedure \cite{zhao2023learning} (Figure~\ref{fig:architecture}). We start with a description of the training only conditional variational autoencoder (CVAE) branch, which encodes the ground truth action chunk together with the proprioceptive state into a latent code that conditions the main decoder. The reconstruction target is truncated to the pose block so that the latent captures only structure shared across the robot and human modalities. The encoder consumes three input streams, each lifted into the hidden dimension $D = 512$ by a dedicated projector, assuming a batch size of $B$: a learnable CLS aggregator broadcast to $(B, 1, D)$; the $K$-step pose projected by a $\mathbb{R}^{K\cdot 57}\to\mathbb{R}^D$ linear projector, and the pose-only action chunk projected per timestep by a linear projector to $H$ tokens of shape $(B, H, D)$. These tokens are concatenated into a length $H + 2$ sequence, summed with a fixed sinusoidal positional embedding, masked along the action axis, and processed by a four-layer post-norm transformer, producing a CLS summary that is linearly projected and split into the mean and log-variance of a $32$-dimensional Gaussian regularized against the unit-Gaussian prior.

The main transformer encoder consumes three token streams: the CVAE latent $z$ (sampled at training and zeroed at inference); the $K$-step poses, projected to one token $(B, 1, 512)$ via a linear projector; and the egoview RGB images, each passed through an ImageNet-pretrained ResNet-18 backbone, mixed by a  $1 \times 1$ convolution. A learnable embedding over the latent and pose tokens and a fixed $2$ dimensional sinusoidal position embedding over the image tokens are added at every layer before passing through a four-layer post-norm transformer encoder with $8$ heads and FFN width $3200$. The single-layer main transformer decoder takes a zero content input of shape $(H, B, 512)$ and a learned positional embedding for DETR-style object queries. The decoder layer applies self-attention, cross-attention onto the encoder memory, and a $512 \to 3200 \to 512$ FFN, followed by a final layer norm. For robot batches, the decoder output is passed through a sibling pair of three-layer MLPs, each $512 \to 512 \to 512 \to d_\text{out}$ with intermediate ReLUs, producing pose ($d_\text{out} = 57$) and qpose predictions ($d_\text{out} = 41$) respectively. For human samples, the qpose branch does not exist, receiving no gradient during human-only pretrain.

In Stage 1, we pretrain the entire architecture with all weights unfrozen using the human data. In Stage 2,  we warm-start all weights except the qpose MLP (cold-started) and finetune on robot data. At inference, all blocks run identically with the trained weights with CVAE style latent set to zero.

\section{Real-World Experiment Details}

\subsection{Detail Evaluation Metrics}
As introduced in Section \ref{subsec:robot_only_ablation}, we implement a normalized metric scoring for each task. For \textbf{T1: Trash Collection}, 0.5 points is awarded if any single piece of trash is successfully deposited into the target basket \textbf{[Single Item]}, and the full 1.0 point is achieved when both pieces are secured \textbf{[Both Items]}. For \textbf{T2: Box Organization}, 0.5 points are awarded when G1 clears the workspace while Vega1 successfully secures the box \textbf{[Clear \& Secure]}, and the remaining 0.5 points are granted if the box is placed on the table without crossing the boundary line, which includes successful recoveries via a corrective nudge from G1 \textbf{[In-Bound Placement]}. For \textbf{T3: Board Tilting}, 0.5 points are awarded if the white board remains on-table and at least one of the three items lands in the basket \textbf{[Partial Containment]}, with the full 1.0 point achieved only when all three items are safely contained \textbf{[Full Containment]}. For \textbf{T4: Doll Passing}, the initial object grasp by Vega1 \textbf{[Initial Grasp]} and the subsequent dual-robot handover \textbf{[Successful Handover]} each contribute 0.5 points. The detailed score distributions and aggregate success rates are summarized in Table \ref{tab:detailed_metric}.

\subsection{Deployment}
During real-world deployment, the centralized inference module monitors the data streams from both the G1 and Vega1. The module takes in real-time egocentric camera images alongside qposes from both robots, which are mapped online via Forward Kinematics to generate synchronized poses. These poses are then transformed into relative head frames with respect to each robot to ensure consistency, according to Section \ref{subsec:data_preprocessing}.
Based on these inputs, the policy predicts target qpose for both robots and base velocity commands for the Vega1 at a fixed frequency of 10 Hz. To guarantee operational safety and preserve multi-robot coordination, we enforce a synchronization protocol: the 10 Hz action commands are only sent if inputs are simultaneously received from both robots. In the event of a asynchronous delay, the system defaults to a zero-order hold, freezing both robots at their last executed actions to prevent uncoordinated movement. Finally, the commanded actions are executed by the hardware via the same low-level control loops utilized during teleoperation (Section \ref{subsec:teleoperation_system_details}), which run independently at 500 Hz to ensure stable, robust, high-frequency action.

\section{Hyperparameters}
\label{sec:hyperparameters}
This section details the hyperparameters used throughout our data collection and training phases. Task-specific clip lengths for robotic and human data are summarized in Tables \ref{tab:robot_parameters} and \ref{tab:human_parameters}, respectively. Additionally, Table \ref{tab:training_parameters} outlines the core training hyperparameters. All models were trained using a single NVIDIA A100-SXM4-80GB GPU per checkpoint. 
\begin{table}[h!]
    \centering
    \caption{Robot Data Recording Parameters}
    \begin{tabular}{lr}
        \toprule
        \textbf{Parameter} & \textbf{Value} \\
        \midrule
        \textbf{T1}: Trash Collection Clip Length ($T$) & $30$ seconds \\
        \textbf{T2}: Box Organization Clip Length ($T$) & $60$ seconds \\
        \textbf{T3}: Board Tilting Clip Length ($T$) & $20$ seconds \\
        \textbf{T4}: Doll Passing Clip Length ($T$) & $30$ seconds \\
        \bottomrule
    \end{tabular}
    \label{tab:robot_parameters}
\end{table}

\begin{table}[h!]
    \centering
    \caption{Human Data Recording Parameters}
    \begin{tabular}{lr}
        \toprule
        \textbf{Parameter} & \textbf{Value} \\
        \midrule
        \textbf{T1}: Trash Collection Clip Length ($T$) & $20$ seconds \\
        \textbf{T2}: Box Organization Clip Length ($T$) & $40$ seconds \\
        \textbf{T3}: Board Tilting Clip Length ($T$) & $13$ seconds \\
        \textbf{T4}: Doll Passing Clip Length ($T$) & $20$ seconds \\
        \bottomrule
    \end{tabular}
    \label{tab:human_parameters}
\end{table}

\begin{table}[h!]
    \centering
    \caption{\textbf{Training Parameters.} $R_\text{role}^\text{G1}$ and $R_\text{role}^\text{Vega1}$ refer to the rotation matrices described in Section~\ref{subsec:data_preprocessing}. $B$ refers to the training batch size, $K$ refers to the history length, and $H$ refers to the future predicted chunk length. Note that $H$ is not to be confused with the executed action chunk length in closed loop. The same parameters are used consistently across the $30$ data robot-only, the $50$ data robot-only, the $60$ data human pretrained and the $30$ robot data finetuned checkpoints.}
    \begin{tabular}{l|c|c|c|c|c}
        \toprule
        Task & $R_\text{role}^\text{G1}$ (Degrees)& $R_\text{role}^\text{Vega1}$ (Degrees)& $B$ & $K$ & $H$ \\
        \midrule
        \textbf{T1} & $(0, 0, 180)$ & $(0, 0, -90)$ & $32$ & $5$ & $400$\\
        \textbf{T2} & $(0, 0, 90)$ & $(0, 0, 180)$ & $32$ & $3$ & $40$ \\
        \textbf{T3} & $(0, 0, -90)$ & $(0, 0, 90)$ & $32$ & $3$ & $40$\\
        \textbf{T4} & $(0, 0, 90)$ & $(0, 0, 180)$ & $32$ & $5$ & $400$ \\
        \bottomrule
    \end{tabular}
    \label{tab:training_parameters}
\end{table}